\documentclass[10pt]{article}

\usepackage{charter} 
\usepackage{fullpage}
\usepackage{titlesec}
\usepackage{amssymb} 
\usepackage{amsfonts}
\usepackage{amsmath} 
\usepackage{cite}
\usepackage{bm} 
\usepackage{graphicx} 
\usepackage{cite} 
\usepackage{authblk} 
\usepackage{algorithmic}
\usepackage{algorithm}
\usepackage{fancyhdr}
\usepackage{float}
\usepackage{url}
\usepackage{mathtools}
\usepackage{multirow}
\usepackage{caption,tabularx,booktabs}
\usepackage{multirow}
\usepackage{diagbox}
\usepackage{threeparttable}

\titleformat*{\section}{\large\bfseries}
\titleformat*{\subsection}{\normalsize\bfseries}
\titleformat*{\subsubsection}{\small\bfseries}

\def\defn{\,\triangleq\,} 
\def\d{\, \mathrm{d}} 
\def\im{\mathrm{j}} 

\def\argmin{\mathop{\mathrm{arg\,min}}} 

\def\uin{{u_{\text{\tiny in}}}}
\def\usc{{u_{\text{\tiny sc}}}}

\def\Real{\textsf{Re}}
\def\Imag{\textsf{Im}}

\newcolumntype{L}[1]{>{\raggedright\arraybackslash}p{#1}}
\newcolumntype{C}[1]{>{\centering\arraybackslash}p{#1}}
\newcolumntype{R}[1]{>{\raggedleft\arraybackslash}p{#1}}

\def\wbf{\mathbf{w}}

\def\ubf{\mathbf{u}}
\def\ubfin{{\mathbf{u}_{\text{\tiny in}}}}
\def\ubfinast{{\mathbf{u}_{\text{\tiny in}}^\ast}}
\def\xbf{\mathbf{x}}

\def\ybf{\mathbf{y}}

\def\ebf{\mathbf{e}}

\def\zbf{\mathbf{z}}
\def\Abf{\mathbf{A}}
\def\Hbf{\mathbf{H}}
\def\Sbf{\mathbf{S}}
\def\Dbf{\mathbf{D}}
\def\Gbf{\mathbf{G}}
\def\Pbf{\mathbf{P}}

\def\Ibf{\mathbf{I}}

\def\xbfhat{\widehat{\mathbf{x}}}

\def\rbm{\bm{r}}

\def\rbmp{{\bm{r}^\prime}}

\def\Rcal{\mathcal{R}}

\def\C{\mathbb{C}}
\def\R{\mathbb{R}}

\def\diag{\textsf{diag}}

\def\Hrm{\textsf{H}}

\usepackage{xspace}
\usepackage{xcolor}

\begin{document}

\title{\Large Efficient and accurate inversion of multiple scattering with deep learning}

\author{\normalsize Yu~Sun$^1$, Zhihao~Xia$^1$, and
		Ulugbek~S.~Kamilov$^{1,2,\ast}$\\
$^1$\emph{Department of Computer Science and Engineering,~Washington University in St.~Louis, MO 63130, USA.}\\
$^2$\emph{Department of Electrical and Systems Engineering,~Washington University in St.~Louis, MO 63130, USA.}\\
$^\ast$\emph{email}: \texttt{kamilov@wustl.edu}
}

\markboth{Deep Learning for Nonlinear Diffractive Imaging}%
{Sun and Kamilov: Deep Learning for Nonlinear Diffractive Imaging}

\date{}
\maketitle 

\begin{abstract}
Image reconstruction under multiple light scattering is crucial in a number of applications such as diffraction tomography. The reconstruction problem is often formulated as a nonconvex optimization, where a nonlinear measurement model is used to account for multiple scattering and regularization is used to enforce prior constraints on the object. In this paper, we propose a powerful alternative to this optimization-based view of image reconstruction by designing and training a deep convolutional neural network that can invert multiple scattered measurements to produce a high-quality image of the refractive index. Our results on both simulated and experimental datasets show that the proposed approach is substantially faster and achieves higher imaging quality compared to the state-of-the-art methods based on optimization.
\end{abstract}


\section{Introduction}
\label{Sec:Introduction}

The problem of reconstructing the spatial distribution of the dielectric permittivity of an unknown object from the measurements of the light it scatters is common in many applications such as tomographic microscopy~\cite{Choi.etal2007} and digital holography~\cite{Brady.etal2009}. The problem is often formulated as a linear inverse problem by adopting scattering models based on the first Born~\cite{Wolf1969} or Rytov~\cite{Devaney1981} approximations. However, these linear approximations are inaccurate when scattering is strong, which leads to reconstruction artifacts for objects that are large or have high permittivity contrasts~\cite{Chen.Stamnes1998}. For strongly scattering objects, it is preferable to use nonlinear measurement models that can account for multiple light scattering inside the object~\cite{Belkebir.Sentenac2003, Belkebir.etal2005, Mudry.etal2012, Tian.Waller2015, Kamilov.etal2015, Zhang.etal2016, Soubies.etal2017, Liu.etal2018, Ma.etal2018}. 

When adopting a nonlinear measurement model, it is common to formulate image reconstruction as an optimization problem. The objective function in the optimization typically includes two terms: a data-fidelity term that ensures that the final image is consistent with measured data, and a regularizer that mitigates the ill-posedness of the problem by promoting solutions with desirable properties~\cite{Ribes.Schmitt2008}. For example, one of the most widely adopted regularizers is total variation (TV), which preserves image edges while promoting smoothness~\cite{Rudin.etal1992}. TV is often interpreted as a sparsity-enforcing $\ell_1$-penalty on the image gradient and has proven to be successful in the context of diffraction tomography with and without multiple scattering~\cite{Sung.Dasari2011, Lim.etal2015, Kamilov.etal2016, Kamilov.etal2016a, Soubies.etal2017, Liu.etal2018, Pham.etal2018}.  

Despite the recent progress in regularized image reconstruction under multiple scattering, the corresponding optimization problem is difficult to solve. The challenging aspects are the nonconvex nature of the objective and the large amount of data that needs to be processed in typical imaging applications. In particular, when the scattering is strong, the problem becomes highly nonconvex, which negatively impacts both the speed of reconstruction and the quality of the final image~\cite{Ma.etal2018}.

In this paper, we consider a fundamentally different approach to the problem of image reconstruction under multiple scattering. Recently, several results have interpreted multiple scattering as a forward-pass of a convolutional neural network (CNN)~\cite{Kamilov.etal2015, Kamilov.etal2016a, Liu.etal2018}. This view inspires us to reconstruct the object by designing another CNN that is specifically trained to invert multiple scattering in a purely data-driven fashion. While our approach is consistent with the recent trend of using deep learning architectures for image reconstruction~\cite{Dong.etal2014, Schmidt.Roth2014, Mousavi.etal2015, Chen.etal2015, Kamilov.Mansour2016, Jin.etal2016, Borgerding.Schniter2016, Han.etal2016, Sinha.etal2017}, it is fundamentally different in the sense that due to multiple scattering our measurement operator is both nonlinear and object dependent (and hence unknown). Our approach is also related to the recent work on reverse photon migration for diffuse optical tomography~\cite{Yoo.etal2017}. However, our focus is on diffractive imaging, where the light propagation is assumed to be deterministic, rather than stochastic as in~\cite{Yoo.etal2017}. Finally, we extensively validated the proposed method on several simulated and real datasets by comparing the method against recent optimization-based approaches based on the Lippmann-Schwinger (LS) model and the TV regularizer~\cite{Soubies.etal2017, Ma.etal2018}. Our results show that it is possible to invert multiple scattering by training a CNN, even when imaging strongly scattering objects for which optimization-based approaches underperform. To the best of our knowledge, the results here are the first to show the potential of deep learning to reconstruct high-quality images from multiple scattered light measurements.

\section{Nonlinear diffractive imaging}
\label{Sec:NonlinearDiffractiveImaging}

In this section, we describe the traditional optimization-based approach for nonlinear diffractive imaging. We first review the image reconstruction and then discuss the details of the physical model for multiple scattering.

\subsection{Nonlinear inverse problem}

We consider an imaging inverse problem
\begin{equation}
\label{Eq:InverseProblem}
\ybf = \Hbf(\xbf) + \ebf \;,
\end{equation}
where the goal is to recover the unknown image ${\xbf \in \R^N}$ from the noisy measurements ${\ybf \in \C^M}$. The measurement operator ${\Hbf: \R^N \rightarrow \C^M}$ models the response of the imaging system and the vector $\ebf \in \C^M$ represents the measurement noise, which is often assumed to be independent and identically distributed (i.i.d.) Gaussian. When the inverse problem is linear, the measurement operator is represented as a measurement matrix $\Hbf \in \C^{M \times N}$.

In practice, problems such as~\eqref{Eq:InverseProblem} are often ill-posed; the standard approach for solving them is by formulating an optimization problem
\begin{equation}
\label{Eq:OptimizationProblem}
\xbfhat = \argmin_{\xbf \in \R^N} \left\{\frac{1}{2}\|\ybf - \Hbf(\xbf)\|_{\ell_2}^2+\Rcal(\xbf)\right\},
\end{equation}
where the data-fidelity term ensures that the final image is consistent with measured data and the regularizer $\Rcal$ promotes solutions with desirable properties. 
Two common regularizers for images include the spatial sparsity-promoting penalty $\Rcal(\xbf) \defn \tau\|\xbf\|_{\ell_1}$ 
and total variation (TV) penalty $\Rcal(\xbf) \defn \tau\|\Dbf\xbf\|_{\ell_1}$, where $\Dbf$ is the discrete gradient operator~\cite{Rudin.etal1992, Candes.etal2006, Donoho2006}. Two common methods for solving optimization problems of form~\eqref{Eq:OptimizationProblem} are FISTA~\cite{Beck.Teboulle2009a} and ADMM~\cite{Afonso.etal2010}, both of which were successfully applied to the problem of image reconstruction from scattered light data~\cite{Zhang.etal2016, Soubies.etal2017, Pham.etal2018, Ma.etal2018}.

\begin{figure}[t]
\begin{center}
\includegraphics[width=5.5cm]{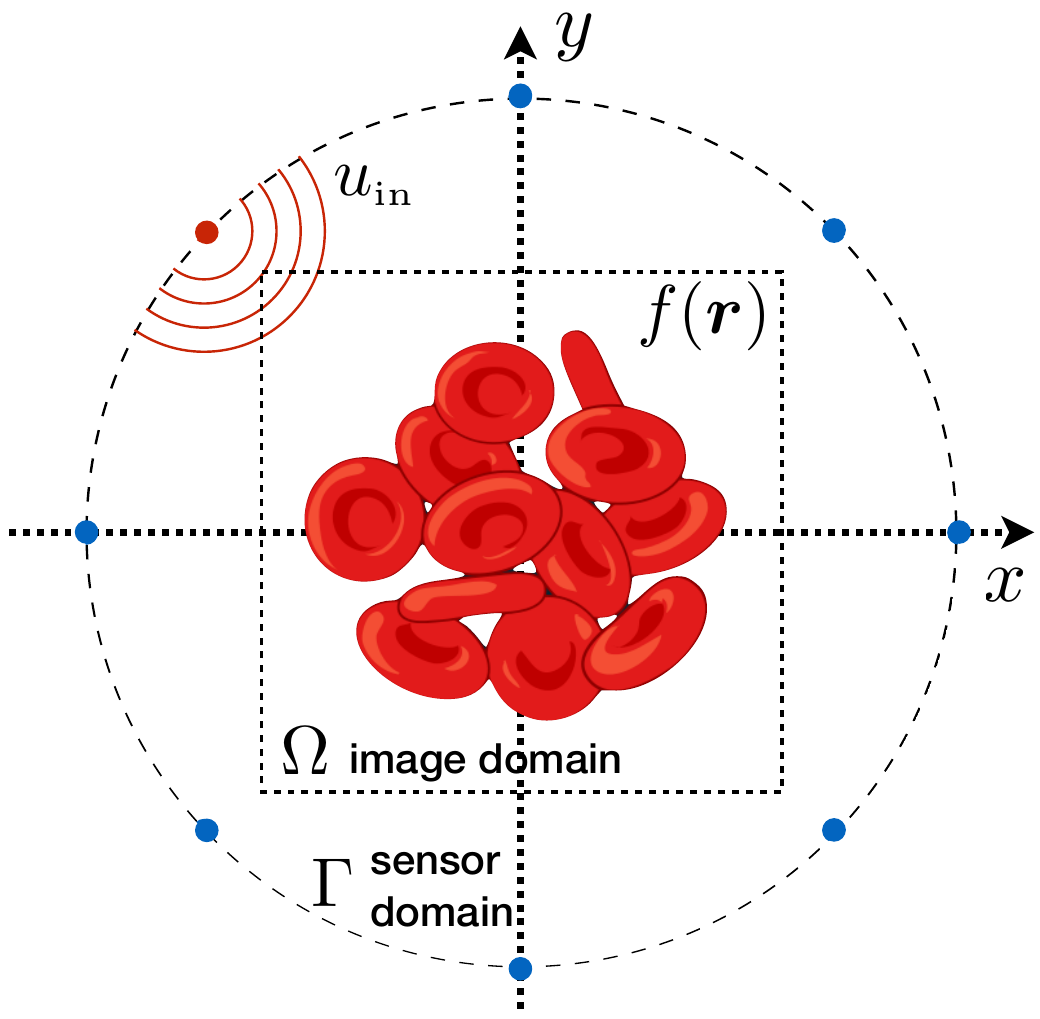}
\end{center}
\caption{Schematic representation of scattering scenarios considered in this letter. An object of a scattering potential $f(\rbm)$ is illuminated with an input wave $\uin$, which interacts with the object and leads to the scattered field $\usc$ at the sensors.}
\label{Fig:Scenario}
\end{figure}

\subsection{Multiple-scattering model}
\label{Sec:ForwardScattering}

Consider the scattering problem illustrated in Fig.~\ref{Fig:Scenario}, where an object of the permittivity distribution $\epsilon(\rbm)$ in the bounded domain ${\Omega \subseteq \R^2}$ is immersed into a background medium of permittivity $\epsilon_b$ and illuminated with the incident electric field $\uin(\rbm)$. We assume that the incident field is monochromatic and coherent, and it is known inside $\Omega$ and at the locations of the sensors. The result of object-light interaction is measured at the location of the sensors as a scattered field $\usc(\rbm)$. The multiple scattering of light can be accurately described by the Lippmann-Schwinger equation~\cite{Born.Wolf2003} inside the image domain
\begin{equation}
\label{Eq:ImageField}
u(\rbm) = \uin(\rbm) + \int_{\Omega} g(\rbm - \rbmp) \, f(\rbmp) \, u(\rbmp) \d \rbmp,\quad(\rbm \in \Omega)
\end{equation}
where $u(\rbm) = \uin(\rbm) + \usc(\rbm)$ is the total electric field,
${f(\rbm) \defn k^2 (\epsilon(\rbm)-\epsilon_b)}$ is the scattering potential, which is assumed to be real, and $k = 2\pi/\lambda$ is the wavenumber in vacuum. The function $g(\rbm)$ is the Green's function, defined as
\begin{equation}
\label{eq:greenfunc}
g(\rbm) \defn \frac{\im}{4} H_0^{(1)}(k_b \|\rbm\|_{\ell_2})
\end{equation}
where $k_b \defn k \sqrt{\epsilon_b}$ is the wavenumber of the background medium and $H_0^{(1)}$ is the zero-order Hankel function of the first kind. Note that the knowledge of the total-field $u$ inside the image domain $\Omega$ enables the prediction of the scattered field at the sensor
\begin{equation}
\label{Eq:SensorField}
\usc(\rbm) = \int_\Omega g(\rbm-\rbmp) \, f(\rbmp) \, u(\rbmp) \d \rbmp.\quad(\rbm \in \Gamma)
\end{equation}

\begin{figure}[t]
\begin{center}
\includegraphics[width=7cm]{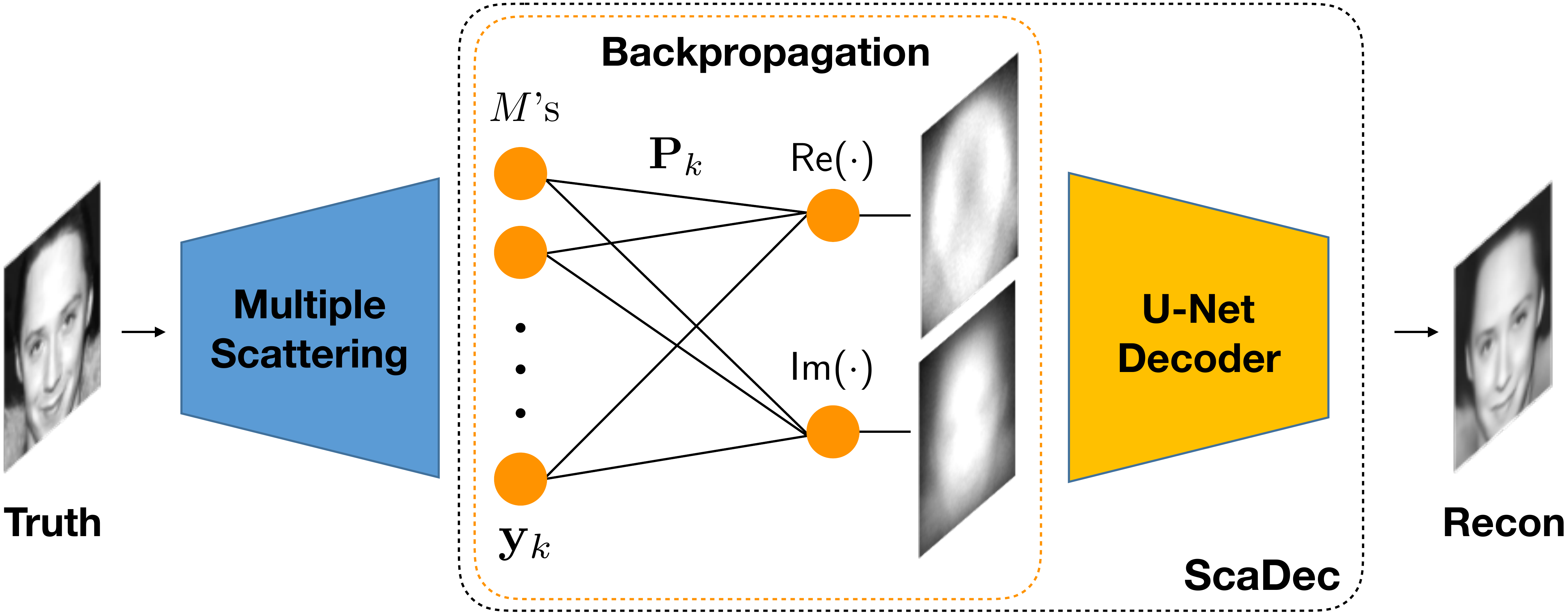}
\end{center}
\caption{The overview of the proposed approach that first backpropagates the data into a complex valued image and then maps this image into the final image with a CNN.}
\label{Fig:ScaDecSchema}
\end{figure}

\begin{figure*}[t]
\begin{center}
\includegraphics[width=13cm]{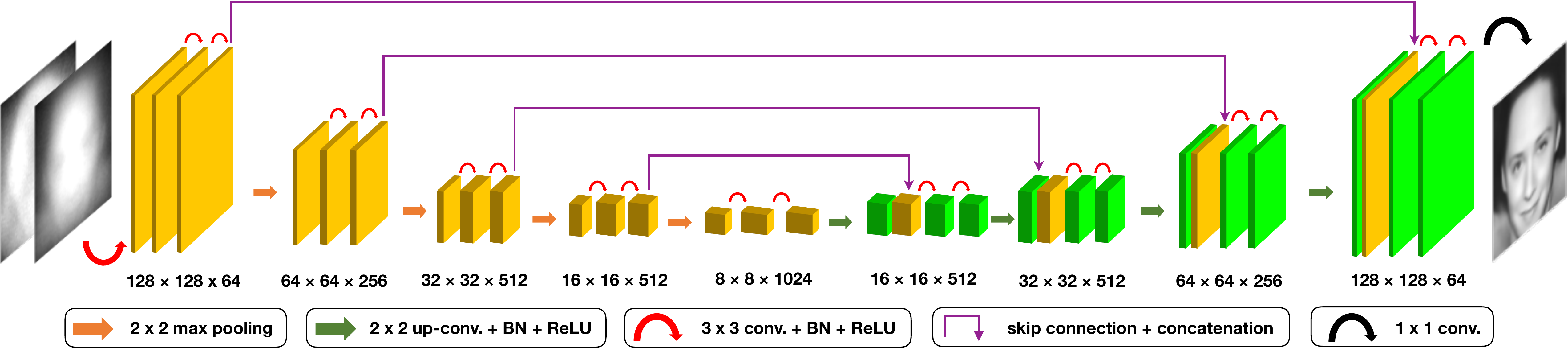}
\end{center}
\caption{Visual illustration of the proposed learning architecture based on U-Net~\cite{RFB15a}. The input consists of two channels for the real and imaginary parts of the backpropagated vector $\wbf \in \C^N$. The output is a single image of the scattering potential $\xbf \in \R^N$.}
\label{Fig:ScaDecArch}
\end{figure*}

The discretization and combination of~\eqref{Eq:ImageField} and~\eqref{Eq:SensorField} leads to the following matrix-vector description of the scattering problem
\begin{subequations}
\label{Eq:NonlinearModel}
\begin{align}
&\ubf = \ubfin + \Gbf (\ubf \cdot \xbf) \label{Eq:ForwardScat2} \\
&\ybf = \Sbf(\ubf \cdot \xbf) + \ebf \label{Eq:ForwardScat1}\;, 
\end{align}
\end{subequations}
where $\xbf \in \R^N$ is the discretized scattering potential $f$, $\ybf \in \C^M$ is the measured scattered field $\usc$ at $\Gamma$, $\ubfin \in \C^N$ is the input field $\uin$ inside $\Omega$, $\Sbf \in \C^{M \times N}$ is the discretization of the Green's function evaluated at $\Gamma$, $\Gbf \in \C^{N \times N}$ is the discretization of the Green's function evaluated inside $\Omega$, $\cdot$ denotes a component-wise multiplication between two vectors, and $\ebf \in \C^M$ models the additive noise at the measurements. Using the shorthand notation $\Abf \defn \Ibf - \Gbf\diag(\xbf)$, where $\Ibf \in \R^{N \times N}$ is the identity matrix and $\diag(\cdot)$ is an operator that forms a diagonal matrix from its argument, we can formally specify the nonlinear forward model as follows
\begin{equation}
\label{Eq:ForwardModel}
\Hbf(\xbf) \defn \Sbf(\ubf(\xbf) \cdot \xbf) \quad\text{where}\quad \ubf(\xbf) \defn \argmin_{\ubf \in \C^N} \left\{\frac{1}{2}\|\Abf \ubf -\ubfin\|_{\ell_2}^2 \right\}
\end{equation}
The recent work has shown that the computation of the operator $\Hbf(\cdot)$ can be interpreted as a CNN and that the gradient of the corresponding data-fidelity term can be efficiently evaluated, enabling efficient optimization~\cite{Kamilov.etal2016a, Soubies.etal2017, Liu.etal2018}. 

\section{Scattering decoder}
\label{Sec:Proposed}

We now describe our proposed deep learning approach called \emph{Scattering Decoder (ScaDec)}.

\subsection{Backpropagation} 

The general framework of our approach is visually illustrated in Fig.~\ref{Fig:ScaDecSchema}.
The first-step in the method is backpropagation, which simply transforms the collected data from the measurement domain to the image domain. We define the backpropagation of the measurements generated by the $k$th transmitter as
\begin{equation}
\label{Eq:BackPropagationSingle}
\zbf_{k} = \Pbf_{k} \ybf_{k} \quad\text{with}\quad\Pbf_{k} \defn \diag(\ubfinast_{,k}) \Sbf^\Hrm,
\end{equation}
where vector $\ybf_{k} \in \C^{M}$ are the measurements consistent with the $k$th transmitter and collected by $M$ receivers, and matrix $\Pbf_{k} \in \C^{N \times M}$ is the backpropagation operator. Inside the operator $\Pbf_{k}$, matrix $\Sbf^\Hrm \in \C^{N \times M}$ is the Hermitian transpose of the discretized Green's function $\Sbf$, and $\ubfinast_{,k}$ is the element-wise conjugate of the incident light field emitted by the $k$th transmitter. The output $\zbf_{k} \in \C^N$ is a complex vector with $N$ elements, which matches the number of pixels in the original image. When the data is collected with multiple transmissions, we define the backpropagation of $K$ transmitters as
\begin{equation}
\label{Eq:BackPropagationMultiple}
\wbf = \sum_{k = 1}^K{\zbf_{k}} = \sum_{k = 1}^K{\Pbf_{k} \ybf_{k}} \;,\\
\end{equation}
where vector $\wbf \in \C^N$ is the linear combination of $\zbf_{k}$ and $K$ denotes the number of transmitters. Note that the backpropagation~\eqref{Eq:BackPropagationMultiple} does not rely on the actual forward model $\Hbf(\cdot)$ in~\eqref{Eq:ForwardModel} which is both nonlinear and object dependent. Remarkably, as we shall see, our simple backpropagation followed by a specific CNN architecture will be sufficient to recover a high-quality image given multiple scattered measurements.

\begin{figure}[t]
\begin{center}
\includegraphics[width=8.5cm]{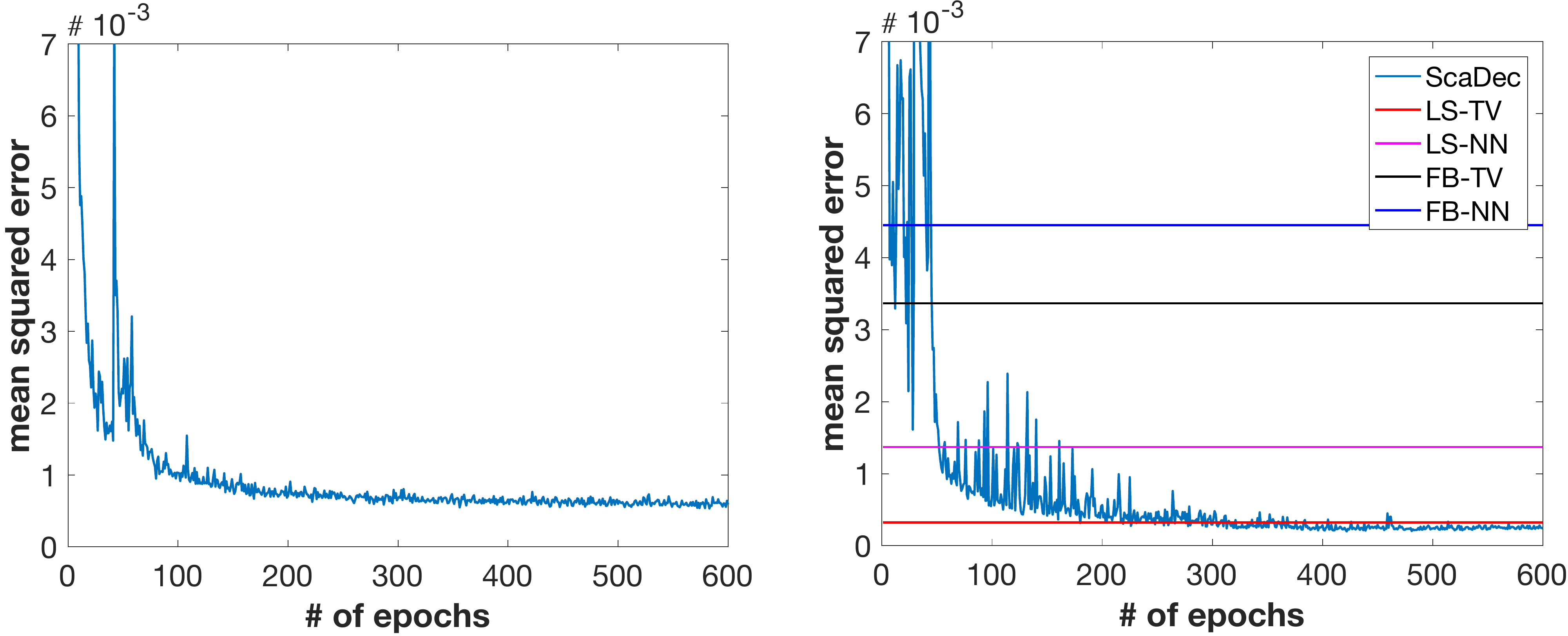}
\end{center}
\caption{Illustration of the convergence of the training  on the dataset of piecewise-smooth objects. The left figure shows the training loss and the right figure shows the validation loss. The horizontal lines on the right show the losses of other algorithms on the same data.}
\label{Fig:Convergence}
\end{figure}

\begin{figure*}[t]
\begin{center}
\includegraphics[width=0.9\textwidth]{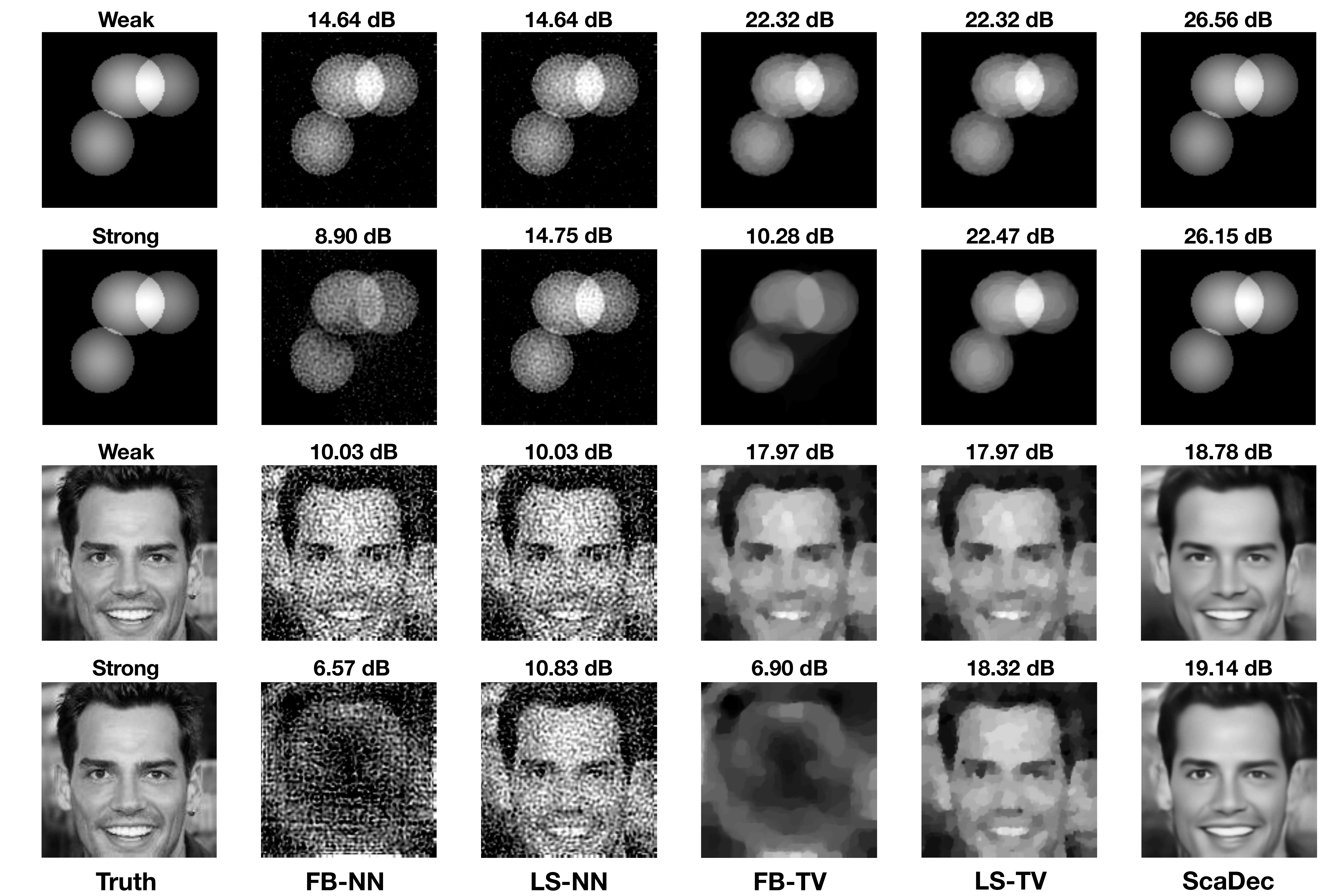}
\end{center}
\caption{Simulated Datasets: Visual comparison of the reconstructed images using the linear model with the first Born approximation regularized by imposing non-negativity~\cite{Sung.etal2009} (FB-NN, column 2) and the total variation~\cite{Sung.Dasari2011} (FB-TV, column 4), the non-linear method in~\cite{Ma.etal2018} regularized by imposing non-negativity (LS-NN, column 3) and the total variation (LS-TV, column 5). The values above images show the SNR (dB) of the reconstruction. The first column shows the true images. Each row corresponds to a different scattering scenario, which is denoted above the leading true image.} 
\label{Fig:VisualExamples}
\end{figure*}

Note that since $\wbf$ is a complex vector, we consider its real and complex parts as two distinct feature maps of the object $f$. Thus, the backpropagation can be viewed as a fixed layer in a CNN with $M$ inputs and two outputs to the subsequent layers (see Fig.~\ref{Fig:ScaDecSchema})~\cite{RFB15a}. The weights inside the layer are characterized by $\Pbf_{k}$'s, and the activation functions for the output nodes are $\Real(\cdot)$ and $\Imag(\cdot)$, respectively.

\subsection{U-Net decoder}
We design the ScaDec~model based on the popular U-Net architecture~\cite{RFB15a}, which was recently applied to various image reconstruction tasks such as X-ray CT~\cite{Jin.etal2016, 2017arXiv170700372Y}. Fig.~\ref{Fig:ScaDecArch} shows a detailed diagram of the proposed CNN architecture. There are two key properties that recommend U-Net for our purpose. 

\begin{enumerate}

\item \emph{Multi-resolution decomposition}: The decoder employs a contraction-expansion structure based on the max-pooling and the up-convolution. This means that given a fixed size convolution kernel ($3 \times 3$ in our case), the effective receptive field of the network increases as the input goes deeper into the network.

\item \emph{Local-global composition}: In each resolution level, the outputs of the convolutional block in the contraction are directly connected and concatenated with the input of the convolutional block in the expansion. The skip connection enables the later layers to reconstruct the feature maps with both the local details and the global texture.

\end{enumerate}
The suitability of U-Net architecture is further corroborated by the results in Section~\ref{Sec:Experiments}, demonstrating the ability of the network to form high-quality images from multiple scattered measurements.

\section{Experimental validation}
\label{Sec:Experiments}

We now present the results of validating our method on simulated and experimental datasets. We evaluate the data-adaptive recovery capability of ScaDec by selecting datasets that consist of images with nontrivial features that can be well represented by a CNN, but are not well captured by fixed regularizers such as TV. The first simulated dataset consists of synthetically generated piecewise-smooth images with sharp edges and smooth Gaussian regions. The second simulated dataset consists of human face images~\cite{liu2015faceattributes}. The experimental dataset is the public dataset provided by the Fresnel Institute~\cite{Geffrin.etal2005}, which consists of experimental microwave measurements of the scattered electric field from 2D targets consisting of foam and plastic cylinders.

\subsection{Results on simulated datasets}

The two simulated datasets were obtained by using a high-fidelity simulation of multiple scattering with the conjugate-gradient solver. Each of the datasets contains 1548 images, separated into 1500 images for training, 24 images for validation, and 24 images for testing. The physical size of images was set to 18 cm $\times$ 18 cm, discretized to a $128 \times 128$ grid. The background medium was assumed to be air with $\epsilon_b = 1$ and the wavelength of the illumination was set to $\lambda = 0.84$ cm. The measurements were collected from 40 transmissions uniformly distributed along a circle of radius $1.6$ m and, for each transmission, 360 measurements around the object were recorded. The simulated scattered data was additionally corrupted by an additive white Gaussian noise corresponding to 20 dB of input signal-to-noise ratio (SNR).

%
%


\begin{table}[t]
\centering
\caption{SNR (dB) comparison of five methods on two datasets}
\label{Tab:RSNRcomparison}
\begin{tabular*}{10cm}{R{1.5cm}C{1.5cm}C{1.2cm}C{0cm}C{1.2cm}C{1.2cm}} \toprule
\multicolumn{1}{r}{\textbf{Method}} & \multicolumn{5}{c}{\textbf{Average SNR over the dataset}}  \\
\cmidrule{2-6}
& \multicolumn{2}{c}{\textbf{Piecewise-smooth}} &  & \multicolumn{2}{c}{\textbf{Human faces}} \\
\cmidrule{2-3} \cmidrule{5-6}
& Weak & Strong &  & Weak & Strong   \\
\cmidrule{1-6}
FB-NN     & 16.49 & 12.79 &  & 10.39 & 6.61 \\
LS-NN  & 16.49 & 16.74 &  & 10.39 & 10.85 \\
FB-TV     & 23.04 & 15.53 &  & 19.79 & 7.08 \\
LS-TV  & 23.04 & 22.57 &  & 19.79 & 20.12 \\
ScaDec    & \textbf{26.14} & \textbf{26.19} &  & \textbf{20.26} & \textbf{20.21} \\ \bottomrule
\end{tabular*}
\end{table}


We evaluated the proposed model in two distinct scenarios associated with the weak and strong scattering. Weak scattering corresponds to the regime where first Born approximation is valid. In particular, we defined the permittivity contrast as $f_\text{\tiny max} \defn (\epsilon_{\text{\tiny max}} - \epsilon_b)/\epsilon_b$, where $\epsilon_{\text{\tiny max}} \defn \max_{\rbm \in \Omega} \{\epsilon(\rbm)\}$. The permittivity contrast quantifies the degree of nonlinearity in the inverse problem, with higher $f_{\text{\tiny max}}$ indicating stronger levels of multiple scattering. We regarded the weakly scattering scenario as $f_{\text{\tiny max}} = 10^{-6}$, whereas the strong scattering scenario was considered as $f_{\text{\tiny max}} = 5 \times 10^{-2}$. For each scenario, we trained a separate ScaDec architecture using the corresponding training dataset with the reconstruction mean squared error (MSE) as the loss function. For quantitively measuring the quality of the reconstructed image $\xbfhat$ with respect to the true image $\xbf$, we used  the signal-to-noise ratio (SNR) defined as
$$\text{SNR}(\xbf, \xbfhat) \defn \max_{a, b \in \R} \left\{10\log_{10}\left(\frac{\|\xbf\|^2_{\ell_2}}{\|\xbf-a\xbfhat+b\|^2_{\ell_2}}\right)\right\},$$
where higher values of SNR correspond to a better match between the true and reconstructed images. As illustrated in Fig.~\ref{Fig:Convergence}, we observed no issues with the convergence of the training for our architecture and datasets. Note that all the SNR and visual results were obtained on a distinct dataset that does not contain images used in training.

Table~\ref{Tab:RSNRcomparison} summarizes the results of comparing ScaDec against the baseline optimization-based methods corresponding to two different priors: nonnegativity constraints on the image and TV. For each prior, we consider the effects of the linearity versus nonlinearity of the measurement model. The linear measurement model is obtained by using the first Born-approximation, while the nonlinear model takes into account multiple scattering by using the full Lippmann-Schwinger equations~\cite{Soubies.etal2017, Ma.etal2018}. Fig.~\ref{Fig:VisualExamples} additionally shows some visual examples of the reconstructed images for each scenario under consideration. Note that the regularization parameters for TV were optimized for the best SNR performance for all the experiments. 

The results confirm that as the level of scattering increases, the performance under the linear inverse problem formulation based on the first Born approximation degenerates with or without regularization. While TV substantially improves the SNR, it also imposes a piecewise-constant structure, leading to blocky artifacts visible in Fig.~\ref{Fig:VisualExamples}. On the other hand, the output of ScaDec substantially outperforms the baseline methods and leads to higher SNR values and to more natural looking images free of blocky artifacts. ScaDec also enjoys good stability in terms of performance, where the reconstruction SNR is nearly identical in weakly and strongly scattering regimes.

Finally, the computational cost of ScaDec is extremely low during the reconstruction stage, where each reconstruction corresponds to a simple forward pass through the CNN. In our case, all the optimization-based methods were run on a pair of 2 CPUs (Intel Xeon processor E5-2620 v4) during testing, while ScaDec was evaluated on a single GPU (NVIDIA GeForce GTX 1080 Ti). We observed that the reconstructing time of ScaDec for a single image was less than 2 seconds in all scenarios, while LS-TV took over 8 and 35 minutes to reconstruct one image in the weakly and strongly scattering cases, respectively. 

\subsection{Results on experimental datasets}

\begin{figure}[t]
\begin{center}
\includegraphics[width=10cm]{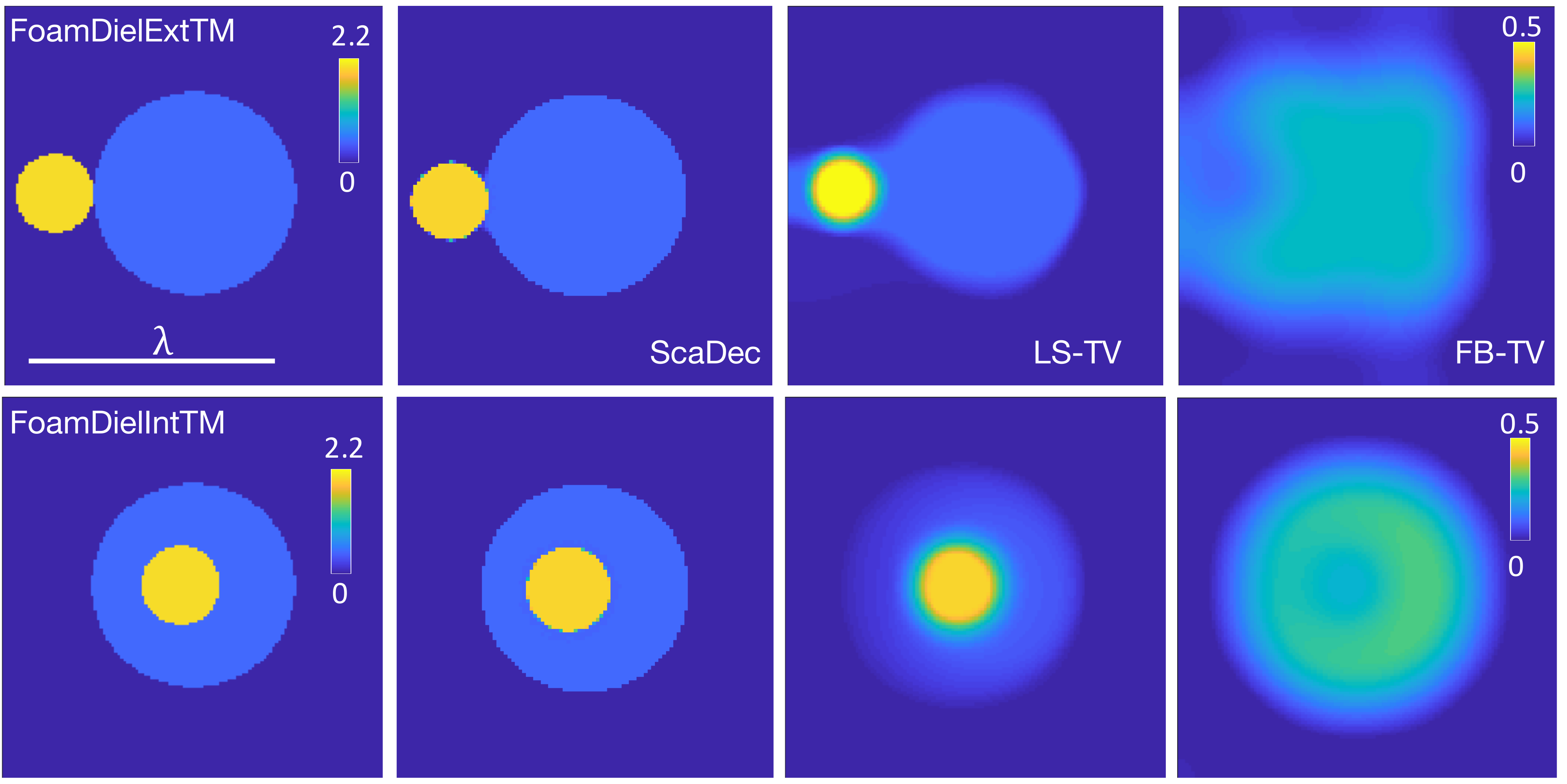}
\end{center}
\caption{Experimental Dataset: Reconstructed images obtained by ScaDec, LS-TV and FB-TV from the data of 2D experimental measurements. The first and second row relate to the setting of \emph{FoamDielExtTM} and \emph{FoamDielIntTM}, respectively. The first column shows the ground truth of each setting. The size of all reconstructed images are $128 \times 128$ pixels. Note that the colormap for FB-TV is different from the rest because the permittivity contrast was extremely underestimated by FB. } 
\label{Fig:ExpExamples}
\end{figure}

For the experimental validation, two 2D settings were considered: \emph{FoamDielExtTM} and \emph{FoamDielIntTM} that consist of a fixed foam cylinder and a plastic cylinder located outside or inside of the foam (Fig.~\ref{Fig:ExpExamples}). In both settings, the objects were placed within a 18 cm $\times$ 18 cm square region, discretized to 128 $\times$ 128 grid, hence, the pixel size of the reconstructed images was 1.4 mm $\times$ 1.4 mm. Total 8 transmitters were uniformly distributed along a circle of radius $1.78$ m, emitting electromagnetic wave towards the objects, and the measurements of the scattered wave were recorded by 360 receivers. Though the dataset contains measurements of a range of wave frequencies, we only consider the case of 5 GHz; hence, the wavelength of the transmission is $\lambda = 60$ mm. The background medium was air with $\epsilon_b = 1$. The permittivity contrasts of foam and plastic were measured as $f_{\text{\tiny max}} = 0.45$ and $f_{\text{\tiny max}} = 2$, respectively~\cite{Geffrin.etal2005}. 

For different settings, we trained the same ScaDec architecture with 6500 pairs of $128 \times 128$ synthetic images and their simulated scattered measurements. The measurements were generated by computing the multiple scattering measurements governed by Lippmann-Schwinger equations. Each image was synthesized with one centered circle with a lower contrast and another randomly-placed circle with a higher contrast. Furthermore, all measurements were corrupted with an additive white Gaussian noise corresponding to 20 dB of input SNR. The ScaDec was trained for 1000 epochs to minimize the MSE between the true image and the restored image.

Visual results of the reconstructed images of different methods are shown in Fig.~\ref{Fig:ExpExamples}, where we compare ScaDec against LS-TV and FB-TV. The first column shows the ground truth of the foam cylinder (light blue) and the plastic cylinder (bright yellow) in each setting. The linear model FB-TV dramatically underestimates the permittivity distribution and fails to reconstruct the shape of objects. On the other hand, the nonlinear model of LS-TV produces better reconstructed images by taking into account both the multiple scattering and the piecewise-constant nature of the image. Finally, the proposed method obtains the highest quality reconstruction in terms of both the contrast value and the shape of objects. The edges of the foam and plastic were clear and sharp, and no obvious degradation of the contrast value was observed. Visually, the results of ScaDec look very close to the ground truth, which is due to the capability of the framework to adapt to the features in the training dataset. Remarkably, the experimental results also illustrate the potential of using simulated data for training, and then deploying the trained CNN for image formation from experimental data.

\section{Conclusion}
We designed and experimentally demonstrated a deep convolutional neural network for solving a multiple scattering problem in diffraction tomography. The proposed method, called ScaDec, successfully reconstructed high-quality images and outperformed state-of-art optimization-based methods in all scenarios. Remarkably, the method trained on simulated data, also succeeded in reconstructing images from real experimental data consisting of highly scattering objects. One of the key advantages of the proposed approach is that the actual process of image formation is substantially accelerated compared to optimization-based reconstruction methods. These features make ScaDec a promising alternative to optimization based methods and opens rich perspectives for efficient correction of scattering in biological samples.

\section*{Acknowledgments}
We gratefully acknowledge the support of NVIDIA Corporation with the donation of the Titan Xp GPU for research.


\bibliographystyle{IEEEtran}


\end{document}